\documentclass[letterpaper, 10 pt, conference]{ieeeconf}  

\IEEEoverridecommandlockouts                              

\usepackage{graphicx}
\usepackage{amsmath}
\usepackage{amssymb}
\usepackage{booktabs}
\usepackage{xcolor}
\usepackage{multirow}
\usepackage{float}
\usepackage{bbding}
\usepackage{subcaption}
\usepackage{hyperref}
\usepackage[noadjust]{cite}

\newcommand{\ie}{\textit{i}.\textit{e}.}
\newcommand{\eg}{\textit{e}.\textit{g}.\ }
\newcommand{\wrt}{w.r.t.\ }
\newcommand{\etc}{\textit{etc}.}
\newcommand{\vs}{\textit{vs}.}

\title{\LARGE \bf
LEF: Late-to-Early Temporal Fusion for LiDAR 3D Object Detection
}

\author{Tong He$^{*}$, Pei Sun, Zhaoqi Leng, Chenxi Liu, Dragomir Anguelov,  Mingxing Tan$^{*}$
\thanks{*Waymo LLC. {\tt\small \{simpleig,tanmingxing\}@waymo.com}}
}

\begin{document}

\maketitle
\thispagestyle{empty}
\pagestyle{empty}

\begin{abstract}
 We propose a late-to-early recurrent feature fusion scheme for 3D object detection using temporal LiDAR point clouds.
 Our main motivation is fusing object-aware latent embeddings into the early stages of a 3D object detector. This feature fusion strategy enables the model to better capture the shapes and poses for challenging objects, compared with learning from raw points directly.
 Our method conducts late-to-early feature fusion in a recurrent manner. This is achieved by enforcing window-based attention blocks upon temporally calibrated and aligned sparse pillar tokens.
 Leveraging bird's eye view foreground pillar segmentation, we reduce the number of sparse history features that our model needs to fuse into its current frame by 10$\times$. 
 We also propose a stochastic-length FrameDrop training technique, which generalizes the model to variable frame lengths at inference for improved performance without retraining.
 We evaluate our method on the widely adopted Waymo Open Dataset and demonstrate improvement on 3D object detection against the baseline model, especially for the challenging category of large objects.
\end{abstract}
\section{Introduction}
\label{sec:intro}

The goal of LiDAR temporal fusion is aggregating learned history information to improve point clouds based tasks. The history information could be of various implicit (\eg latent embeddings), explicit (\eg point clouds, 3D box tracklets) representations or a mixture of both, depending on the models and tasks at hand.
Temporal fusion is critical for multiple driving related tasks, such as 3D object detection,  tracking, segmentation, and behavior prediction. Here we mainly study LiDAR-based fusion methods for 3D object detection, which is a crucial task for recognizing and localizing surrounding objects in modern autonomous driving systems.
Point clouds of a single frame can only serve as partial observation of the scenes, lacking complete coverage of environment context and agent dynamics. This information bottleneck is caused by several factors such as object self-occlusion, occlusion by other objects, sensor field-of-view limitation, and data noises. Moreover, for moving objects, models with only single-frame data will struggle to understand their short-term states (velocities, accelerations) and long-term intentions (future trajectories). Tackling these issues demands effective ways of LiDAR temporal fusion, which can enable the model to understand scene / object attributes and dynamics from a wide time horizon.

\begin{figure}[t]
\captionsetup[sub]{font=small}
\centering
\begin{subfigure}[b]{\linewidth}
\includegraphics[width=\columnwidth]{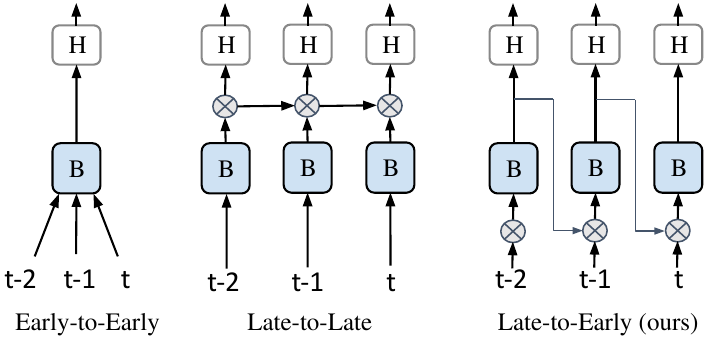}
\caption{Overview structures of three temporal fusion approaches, where \texttt{B} denotes the backbone, \texttt{H} denotes the detection head.}
\label{fig:teaser_strategy}
\vspace{+4 pt}
\end{subfigure}
\begin{subfigure}[b]{\linewidth}
\includegraphics[width=\columnwidth]{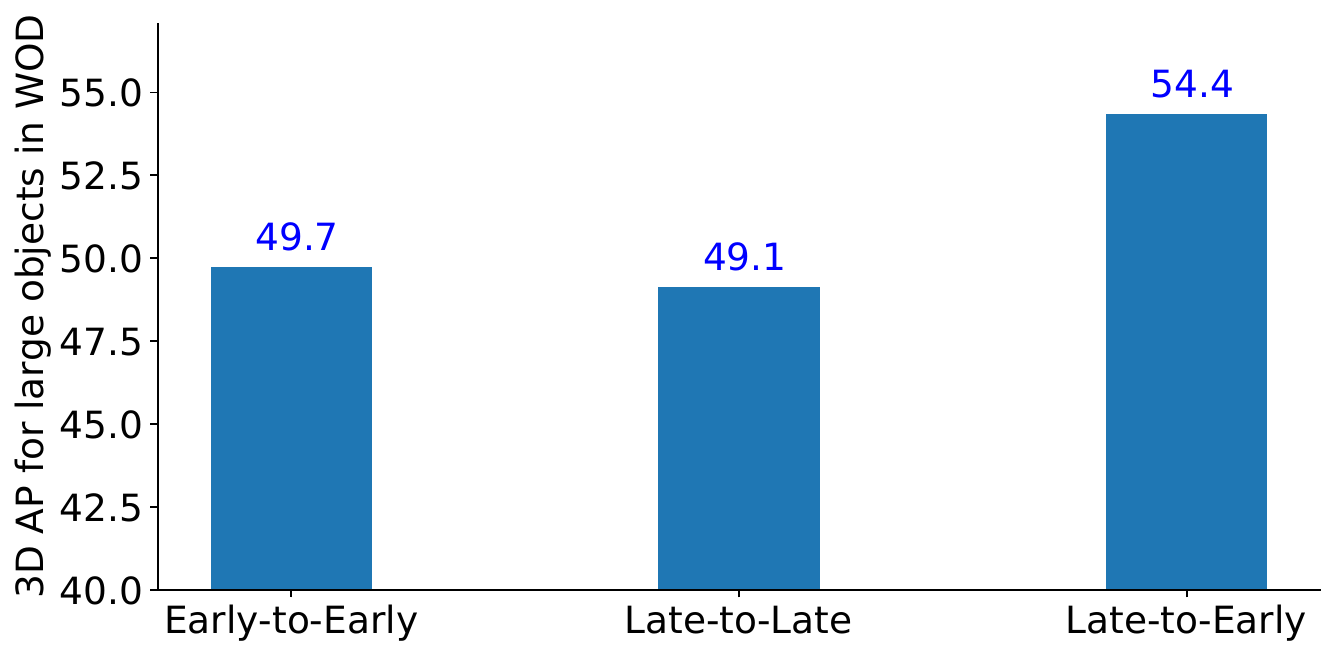}
\caption{Performance comparisons on Waymo Open Dataset.}
\label{fig:teaser_performance}
\end{subfigure}
\caption{\textbf{Comparisons of temporal fusion approaches}. Our late-to-early fusion approach achieves better detection quality (\eg 54.4 3D AP for the challenging large objects) than previous early-to-early and late-to-late methods.}
\label{fig:teaser}
\vspace{-20 pt}
\end{figure}

The main challenge of temporal fusion is how to represent and aggregate the long-sequence information of history frames.
See \autoref{fig:teaser_strategy} for a high-level illustration and comparison.
Generally speaking, previous solutions can be classified into two types.
One of the most widely used methods is early-to-early fusion based point cloud stacking. Multi-frame LiDAR points are directly stacked together as model inputs, resulting in better performance than a single frame of LiDAR points. However, the performance quickly saturates when more frames are simply stacked together \cite{huang2020lstm} without careful modeling of the inter-frame relationships. Moreover, each frame needs to be repeatedly processed when they are stacked into different adjacent frames, greatly increasing computation cost. Fitting long sequences will also greatly increase memory cost, reduce model efficiency or even result in out of memory (OOM) issues. Ideally, a model should leverage what it has already learned from the data, not simply stacking its raw sensory inputs.
To overcome this issue, another type of fusion methods turn to late-to-late fusion so as to utilize the learned history embeddings. A representative method is ConvLSTM \cite{huang2020lstm} which recurrently fuses latent embeddings between consecutive frames at deep layers of the model. This approach reduces memory usage and computation cost, but its results are usually inferior to early-to-early fusion, as shown in  \autoref{fig:teaser_performance}. We suspect that this is because the backbone only has access to single-frame data before late fusion happens. The task of understanding temporally fused deep features falls upon the detection heads, which usually consist of low-capacity multi-layer perceptron (MLP) layers. Consequently, most state-of-the-art LiDAR 3D object detectors (\eg PVRCNN++ \cite{shi2020pv,Shi2021PVRCNNPF}, CenterPoint \cite{yin2021center}, SST \cite{fan2022embracing}, SWFormer \cite{sun2022swformer}, \etc) still rely on early-to-early fusion with point cloud stacking.

In this paper, we propose a new fusion method named \textbf{LEF}: \textbf{L}ate-to-\textbf{E}arly temporal \textbf{F}usion.
We argue that this fusion scheme can leverage learned history knowledge, and in the meantime
its backbone does not suffer from single-frame data deficiency issues.
Long history LIDAR fusion is a fundamental block for autonomous driving, and our work opens a promising direction to achieving that goal.
There are three main contributions in our paper:
\begin{itemize}
    \item We propose a recurrent architecture that fuses late-stage sparse pillar features into early stages of the next frame. To align the underlying static objects, we propose an inverse calibration and alignment module to fuse history and current sparse sets of pillar features. As for moving objects, we leverage window-based attention layers, which can associate relevant features within the windows and thus connect pillar tokens that belong to the same object. 
    \item While point stacking struggles to cache and preprocess huge point clouds as history length grows, we leverage a bird's eye view (BEV) foreground pillar segmentation module to achieve long-sequence fusion at a low constant cost. The number of sparse voxels that our model needs to fuse at each recurrent step can be reduced by over 10$\times$ via the foreground segmentation process.
    \item We also propose a stochastic-length FrameDrop training recipe. It exposes the model to an augmented large motion space of pillar trajectories across time. Thus our recurrent model can capture different speed objects, and generalize to variable frame lengths during inference for improved performance.

\end{itemize}

The proposed late-to-early temporal fusion scheme leads to improved 3D detection results on the widely used Waymo Open Dataset (WOD)~\cite{sun2020scalability} and demonstrates large gains on challenging large objects. We also conduct extensive ablation studies on various design choices made in our method, providing several interesting insights.
\section{Related Work}
\label{sec:related_work}

\textbf{3D Object Detection}.
LiDAR-based 3D object detection plays an essential role in autonomous driving. 
Early efforts of research such as PointRCNN \cite{Shi2019PointRCNN3O} usually operate on raw 3D point clouds through PointNet(++) \cite{Qi2018FrustumPF,Qi2017PointNetDL, Qi2017PointNetDH}. But they struggle to generalize to large-scale data, such as long-sequence fused LiDAR \cite{sun2020scalability} with millions of points. Heavily relying on MLP-based backbones, these detectors are soon outperformed by models with more advanced architectures like submanifold sparse convolution \cite{graham2017submanifold} or Transformers \cite{vaswani2017attention,zhao2021point,mao2021voxel}.
By voxelizing free-shape point sets into regular 2D\footnote{2D-shape voxels are often referred to as pillars.} or 3D-shape voxels, LiDAR-based detectors \cite{lang2019pointpillars, Yan2018SECONDSE, zhou2018voxelnet} can leverage numerous advancements on image 2D object detection, and start to demonstrate promising 3D detection results. Particularly, CenterPoint \cite{yin2021center} utilizes sparse convolution layers and CenterNet-based detection heads~\cite{Zhou2019ObjectsAP} to predict 3D boxes. Some recent works, such as SST \cite{Fan2022EmbracingSS} and SWFormer \cite{sun2022swformer}, exploit Swin-Transformer \cite{Liu2021SwinTH} and push the detection performance to a new state of the art.
Meanwhile, several methods \cite{shi2020pv, Shi2021PVRCNNPF, sun2021rsn, Meyer2019LaserNetAE, Ngiam2019StarNetTC, wang2020pillar, Chai2021ToTP, Fan2021RangeDetID, li2021lidar, sheng2021improving, liu2022lidarnas} look into alternative LiDAR representations and strive towards a balance between detection efficiency and efficacy. 

\textbf{LiDAR Temporal Fusion}.
Compared with the rapid progresses achieved on 3D detection backbones, approaches of LiDAR temporal fusion are less well-studied. Point clouds of a single frame in WOD~\cite{sun2020scalability} have already caused huge computation burden (\ie, $\sim$200\textit{k} points), let alone long history sequences. As briefly discussed in the introduction section, LiDAR temporal fusion solutions can be generally classified into three types: 
early-to-early, late-to-late and late-to-early fusion. Early-to-early fusion is also referred to as point cloud stacking. It is most widely adopted in recent LiDAR object detectors (\eg CenterPoint \cite{yin2021center}, RSN \cite{sun2021rsn}, SWFormer \cite{sun2022swformer}, \etc) due to its simple setup. Multi-frame point sets are merged together. Timestamp offsets \wrt to the current frame are appended to sensory signals of each 3D point to serve as markers indicating different frame sources. However, point stacking struggles to work on long sequences due to the cost of fusing, saving and jointly preprocessing millions of points. It is also possible to use a Transformer to early fuse point clouds from different frames~\cite{yuan2021temporal}. While early-to-early fusion simply stacks raw sensory inputs without carefully modeling inter-frame relationships and ignores knowledge learned from prior frames, late-to-late fusion tries to tackle these issues by ConvLSTM \cite{huang2020lstm, yin2020lidar}. It recurrently fuses sparse latent embeddings between deep layers of the backbone with improved efficiency than point stacking, but the results are often not as competitive as early-to-early fusion. This is presumably because its backbone can only utilize single-frame data until fusion happens at deep layers. 3D-MAN~\cite{yang20213d} may also be viewed as a form of late-to-late fusion, because the temporal fusion in this method is done through various kinds of cross-attention between box proposals and features in the memory bank, which are both after the backbone of its network. FaF~\cite{luo2018fast} studied both early fusion and late fusion. To the best of our knowledge, late-to-early fusion has not been explored before in LiDAR detectors. A similar fusion framework is studied in \cite{li2022bevformer} but targeting on camera-based detection. It faces very different challenges from our problems. We need to process sparsely distributed 3D data at wide ranges, which requires dedicated designs for sparse features alignment, fusion and also new training recipes.

Finally, we note that our review so far concentrates on a single-stage trainable model that internalizes the temporal fusion schemes. It is also possible to follow up the box predictions with a second-stage offline refinement, using the terminology from a recent exemplar of this two-stage approach, MPPNet~\cite{chen2022mppnet}. MPPNet runs a pre-trained CenterPoint \cite{yin2021center} on 4-frame stacked LiDAR point clouds to generate anchor boxes, which will then be tracked and aggregated across long sequences. Specifically, latent embeddings or raw points within the box regions of one frame will be cropped and intertwined with those extracted from other frames in order to refine the box states. The key differentiating factor about the two-stage approach is that the two stages / models are trained separately~\cite{chen2022mppnet}, suggesting that the improvement inherently built into the first stage, like ours, is complementary to the second-stage innovation.
\section{Method}
\label{sec:method}

\subsection{Problem Statement}
We use $\{P_{i}\}$, $i=1,...,T$ to represent a consecutive sequence of LiDAR point clouds with $P_i:\{X_{i,j} \in {\mathbb R}^3\}$, $j=1,...,N_{i}$. Our goal is to detect 3D object boxes $\{B_{i,m}\}$, $m=1,...,M_{i}$ for each frame-$t$ using $\{P_{i}\ |\ i \leqslant t\}$. Ideally the model should be capable of fusing history information $F(P_1,...,P_{t})$ up to the current timestamp-$t$, where $F(\cdot)$ denotes the fusion function.
LiDAR temporal fusion is known to be an open challenge due to the sparse and wide-range spatial distribution of point clouds, let alone diverse object dynamics. Currently early-to-early fusion (\ie, point stacking) is most widely used $P_{t-l}\cup ... \cup P_{t}$, which is easy to implement. However, due to memory constraint the sequence length is usually small, e.g. $l \in \{2,3\}$. Moreover, point clouds $\{X_{i,j}\}$ of one frame have to be repeatedly processed for $(l+1)$ times when we conduct model inference on adjacent frames, causing huge waste of computation. 
As for detection performance, whether directly stacking the raw sensory inputs without reusing learned history knowledge can lead to the optimal results also remains questionable.

\begin{figure*}[t]
\centering
\includegraphics[width=15.5cm]{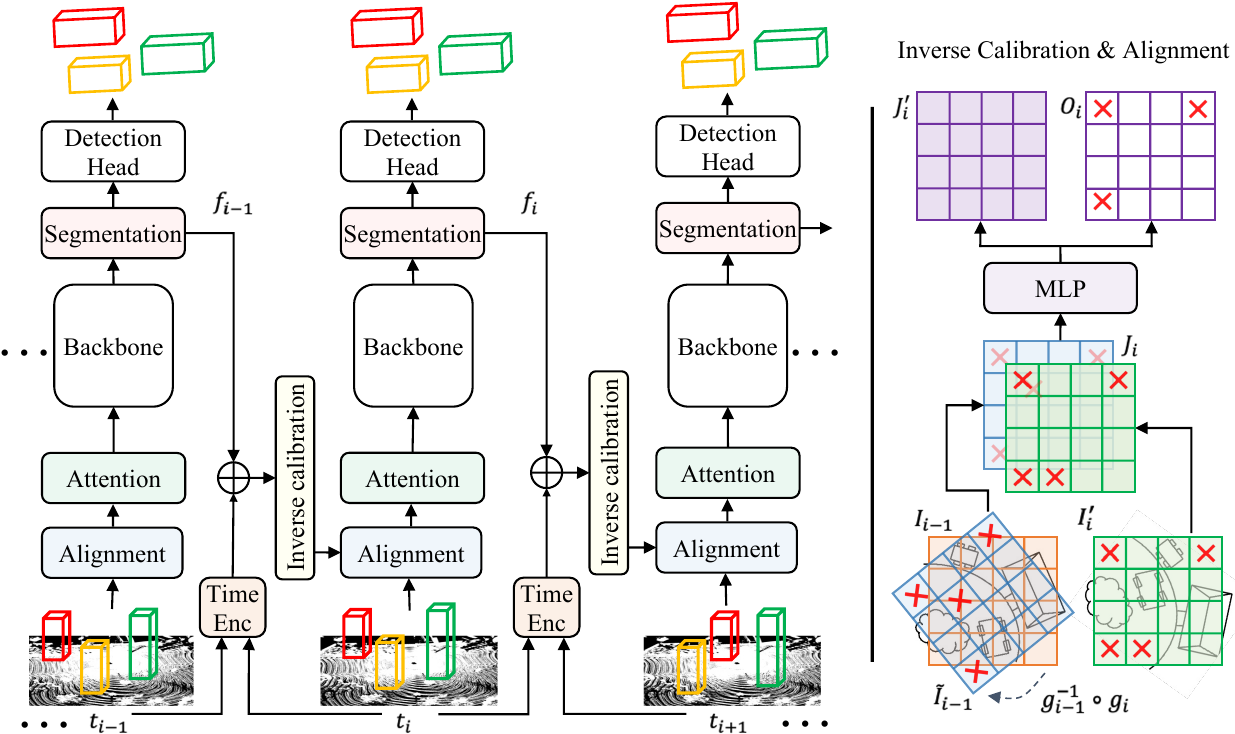}
\caption{\textbf{Detection pipeline with our proposed LEF}. In each forward pass, the early-stage pillar encoding will be aligned and fused with the history late-stage foreground pillar features $f_{i-1}$. The alignment is achieved by an inverse calibration and alignment process (Section \ref{sec:inverse_calib_and_align}) that enables pillar features of the underlying static objects to be matched. To effectively associate moving object features, we further use window-based attention blocks (Section \ref{sec:method_window_attention}) to connect relevant pillars. Outputs from the attention fusion layers will then be fed into the main backbone network (\eg SWFormer~\cite{sun2022swformer}), followed by a foreground pillar segmentation layer and the final detection head~\cite{yin2021center} for 3D bounding box predictions.}
\label{fig:pipeline}
\vspace{-10 pt}
\end{figure*}

\subsection{Recurrent Late-to-Early Fusion}
To address the aforementioned issues, we propose a recurrent late-to-early temporal fusion strategy. As shown in \autoref{fig:pipeline}, the fusion pipeline works like a ``Markov chain'', which can accumulate history information from long sequences and reduce redundant computation. Thus, the fusion function $F(\cdot)$ can be iteratively defined as:

\vspace{-3 pt}
\begin{align} \label{eq:l2e_recurrent_fusion}
\begin{split}
f_{i} = \psi(h(f_{i-1} \oplus \tau (t_{i} - t_{i-1}), \nu (\{X_{i,j}\})))
\end{split}
\end{align}

\noindent where $f_{i-1}$ indicates history deep-layer voxel embeddings, and $\tau (\cdot)$ is a Sinusoidal function for encoding the timestamp offset. $\nu(\cdot)$ represents VoxelNet \cite{zhou2018voxelnet} used to obtain pillar features from point clouds. $h(\cdot)$ is the backbone for recurrent fusion and multi-scale sparse pillar features extraction, and $\psi(\cdot)$ is the foreground segmentation module.

\textbf{History features}. Particularly, we use the latent features of segmented foreground pillars as $f_{i-1}$ and pass them into the next timestamp. Without loss of generality, we use SWFormer \cite{sun2022swformer} as our backbone and center-based detection heads \cite{yin2021center} as examples in our following discussion if needed. The diagram is plotted in \autoref{fig:pipeline}. The model works on sparse pillar tokens and thus the segmentation outputs can be written as $f_{i-1}:\{V_{i-1,k} \in {\mathbb R}^{2+d} \}$, $k=1,...,K_{i-1}$. The first two dimensions record BEV coordinates of the pillars and the rest are extracted embeddings (\ie, $d=128$), which contain rich scene and object-aware information. Moreover, compared with the raw point clouds size $N_{i-1}$ ($\sim$200$k$), the foreground pillar feature set size $K_{i-1}$ ($\sim$2$k$) is much smaller. Therefore, we are motivated to fuse these deep-layer features into early stages of the next frame in order to efficiently reuse learned high-level knowledge for 3D detection, especially on challenging large objects.

\textbf{Fusion location}. To achieve recurrent late-to-\textit{early} fusion, we fuse $f_{i-1}$ with VoxelNet \cite{zhou2018voxelnet} outputs $\nu (\{X_{i,j}\}) \mapsto \{V^{'}_{i,n} \in {\mathbb R}^{2+d} \}$, $n=1,...,N^{'}_{i}$ before they are fed into the the main backbone network. Meanwhile, instead of early fusion before the backbone, some may argue that an alternative way is conducting late fusion after the backbone process, which is close to the network stage where $f_{i-1}$ is extracted. Diagrams of these two different fusion locations are plotted in \autoref{fig:teaser}. We think that presumably late fusion can cause the backbone $\texttt{B}$ to lose access to temporally aggregated LiDAR sequence information, and thus the low-capacity detection heads $\texttt{H}$ will struggle to understand fused features and predict object poses and shapes. Ablation studies on early-to-early, late-to-late and our proposed late-to-early fusion methods are provided in \autoref{tab:fusion_type} and Section \ref{sec:ablation_studies}, which empirically proved the advantages of our approach.

\subsection{Inverse Calibration and Alignment}
\label{sec:inverse_calib_and_align}
While image sequences are naturally aligned across different frames by the shapes (height, width, channel), sparse sets of pillar features $\{V_{i-1,k}\}$, $\{V^{'}_{i,n}\}$ are neither aligned nor with the same cardinality (\ie, $K_{i-1} \neq N^{'}_{i}$).
Intuitively one could convert sparse features into dense BEV maps $\{V_{i-1,k}\} \mapsto I_{i-1} \in {\mathbb R}^{H \times W \times d}$, $\{V^{'}_{i,n}\} \mapsto I^{'}_{i} \in {\mathbb R}^{H \times W \times d}$ and then align them. However, as \autoref{fig:pipeline} shows, directly doing so without proper calibration can result in misalignment between underlying objects of the scene. This is because pillar features extracted by the backbones are from their corresponding local vehicle coordinates with poses of $g_{i-1} \in {\mathbb R}^{4 \times 4}$, $g_i \in {\mathbb R}^{4 \times 4}$. To alleviate this misalignment issue, we need to calibrate the history BEV maps $I_{i-1}$.

\begin{align} \label{eq:history_calibration}
\begin{split}
I_{i-1} \circ g^{-1}_{i-1} \circ g_{i} \mapsto \tilde{I}_{i-1}
\end{split}
\end{align}

\noindent here $\circ$ means applying vehicle coordinates transformation and $\tilde{I}_{i-1}$ represents the calibrated BEV maps.

However, in practice if we apply forward calibration upon $I_{i-1}$ we might get more than one pillars that fall into the same discrete coordinates within $\tilde{I}_{i-1}$. To address this issue we conduct inverse transformation from $\tilde{I}_{i-1}$ to $I_{i-1}$ and sample the history BEV features. We use zero padding to fill in the pillar features of empty samples and also for out-of-view locations, \eg red cross markers in \autoref{fig:pipeline}.
The inversely calibrated history maps now can be aligned with current maps by feature concatenation $\tilde{I}_{i-1} \oplus I^{'}_{i} \mapsto J_{i} \in {\mathbb R}^{H \times W \times 2d} $. Next, we apply a MLP on $J_{i}$ for dimension reduction (\ie, $2d \mapsto d$) and get the temporally aligned pillar features $J^{'}_{i}$. Note that not all the coordinates within $J^{'}_{i}$ have valid features. We use the union BEV boolean mask $O_{i} \in {\mathbb R}^{H \times W}$ obtained from the current and calibrated history BEV features to mark valid coordinates of $J^{'}_{i}$. Thus, we do not lose the data sparsity.

\subsection{Window-based Attention Fusion}
\label{sec:method_window_attention}
Pillars of the static objects are effectively aligned after the prior steps, but the moving ones are still facing the misalignment issue. One solution is to apply flow estimation to further calibration the history BEV features $\tilde{I}_{i-1}$ before temporal alignment with $I^{'}_{i}$. But that requires adding additional occupancy flow models, losses and feature coordinates transformation, which might greatly increase the computation overhead of the 3D object detector. Therefore, we propose to learn such association implicitly from the data by window-based attention blocks. We sparsify the dense BEV feature map $J^{'}_{i}$ and its boolean mask $O_{i}$ into a sparse set of pillar tokens $\{V^{''}_{i,u}\}$, $u=1,...,U_{i}$. Usually we have $U_i \geqslant N^{'}_{i}$. Because the cardinality $U_i$ means the number of fused pillars after temporal alignment between the history and current features through the steps in Section \ref{sec:inverse_calib_and_align}.
While $\{V^{''}_{i,u}\}$ is used as the query tensor for the attention blocks, we can make different choices when determining the key and value tensors: using $\{V^{''}_{i,u}\}$ again or the sparsified set of history pillar tokens in (\ref{eq:history_calibration}):
$\tilde{I}_{i-1} \mapsto \{ \tilde{V}_{i-1,c} \}$, $c=1,...,\tilde{K}_{i-1}$. Most often, $\tilde{K}_{i-1} \leqslant K_{i-1}$ due to out-of-view truncation after vehicle coordinates calibration.

The resulting variants are: self / cross / mix-attention. In self-attention the key and value tensors are the same as query. Cross-attention uses $\{ \tilde{V}_{i-1,c} \}$ as key and value and mix-attention uses the union set of prior two attention variants. We apply Sinusoidal functions based absolute positional encoding to inform the attention blocks of the sparse pillar coordinates within a window. Detailed ablation studies on different attention designs are provided in Section \ref{sec:ablation_studies}. With window-based attention fusion, features of both static and moving pillars now can be associated and fused for later being passed into the main backbone network.

\begin{table*}[t]
\begin{center}
\caption{\textbf{Overall performance comparisons on Waymo Open Dataset}.
Refine means that the detectors need an additional step of box refinement via feature pooling and fusion from the box areas, which usually increases time cost and might not be end-to-end trainable. For fair comparisons we focus on single-stage detectors without (w/o) box refinement.}
\resizebox{0.77\linewidth}{!}{
\begin{tabular}{l|c||cc||cc}
\toprule
\multirow{2}{*}{Method}       & \multirow{2}{*}{Refine} & \multicolumn{2}{c||}{Test set 3D AP/APH} & \multicolumn{2}{c}{Validation set 3D AP/APH}         \\
  &  & L1          & L2          &  L1 &  L2\\ \midrule
3D-MAN~\cite{yang20213d}  & with    & \multicolumn{1}{c|}{78.71 / 78.28} & 70.37 / 69.98 &  \multicolumn{1}{c|}{74.53 / 74.03} & 67.61 / 67.14\\
CenterPoint \cite{yin2021center}  & with    & \multicolumn{1}{c|}{80.20 / 79.70} & 72.20 / 71.80 &  \multicolumn{1}{c|}{76.60 / 76.10} & 68.90 / 68.40\\
SST \cite{fan2022embracing}         & with    & \multicolumn{1}{c|}{80.99 / 80.62} & 73.08 / 72.72 & \multicolumn{1}{c|}{77.00 / 76.60} & 68.50 / 68.10 \\
PVRCNN++ \cite{shi2020pv}    & with    & \multicolumn{1}{c|}{81.62 / 81.20} & 73.86 / 73.47 & \multicolumn{1}{c|}{79.30 / 78.80} & 70.60 / 70.20 \\
MPPNet \cite{chen2022mppnet}    & with    & \multicolumn{1}{c|}{84.27 / 83.88} & 77.29 / 76.91 & \multicolumn{1}{c|}{82.74 / 82.28} & 75.41 / 74.96 \\
CenterFormer \cite{zhou2022centerformer}    & with    & \multicolumn{1}{c|}{84.70 / 84.40} & 78.10 / 77.70 & \multicolumn{1}{c|}{78.80 / 78.30} & 74.30 / 73.80 \\ \midrule
PointPillars \cite{lang2019pointpillars} & w/o     & \multicolumn{1}{c|}{68.60 / 68.10} & 60.50 / 60.10  & \multicolumn{1}{c|}{63.30 / 62.70} & 55.20 / 54.70 \\
RSN \cite{sun2021rsn}         &  w/o      & \multicolumn{1}{c|}{80.70 / 80.30} & 71.90 / 71.60 & \multicolumn{1}{c|}{78.40 / 78.10} & 69.50 / 69.10 \\
SWFormer \cite{sun2022swformer} &  w/o      & \multicolumn{1}{c|}{82.25 / 81.87} & 74.23 / 73.87 & \multicolumn{1}{c|}{79.03 / 78.55} & 70.55 / 70.11 \\
\textbf{LEF (ours)}        &  w/o      & \multicolumn{1}{c|}{\textbf{83.39} / \textbf{83.02}} & \textbf{75.51} / \textbf{75.16} & \multicolumn{1}{c|}{\textbf{79.64} / \textbf{79.18}} & \textbf{71.37} / \textbf{70.94} \\ \bottomrule
\end{tabular}%
}
\label{tab:main_results_test}
\end{center}
\vspace{-13 pt}
\end{table*}

\subsection{Stochastic-Length FrameDrop}
To enable robust training upon long sequences, we randomly drop history frames from $(P_1,...,P_{t})$ during each training iteration. In other words, we randomly sample $S_{i}$ history frames, with $S_{i}$ being a stochastic number at different training steps and the sampled frames are not necessarily adjacent ones. In comparison, the previous LiDAR temporal fusion methods usually fix $S_{i}$ to be a constant (\eg 3 or 4) and sample consecutive frames. We apply stop gradient between each recurrent pass when fusing deep-layer history features into early layers of the next frame, without which long-sequence training of 3D object detectors can easily get intractable or run into OOM.
During training, the model only predicts 3D boxes $\{\hat{B}_{i,m}\}$ in the last forward pass. Losses are enforced upon certain intermediate outputs (\eg foreground pillar segmentation) and the final box parameter predictions (\eg shapes and poses).

\vspace{-0.15in}
\begin{align} \label{eq:total_loss}
\begin{split}
L = \lambda_{1}L_{seg} + \lambda_{2}L_{center} + L_{box}
\end{split}
\end{align}

\noindent in which $L$ means the total losses. $L_{seg}$ is focal loss for foreground segmentation. $L_{center}$ is also based on focal loss but for object-center heatmap estimation \cite{zhou2019objects, yin2021center}. $L_{box}$ contains SmoothL1 losses for box azimuth, center offsets and sizes regression. A detailed explanation is in \cite{sun2022swformer}.

The training randomness introduced in LiDAR sequence sampling enables the model to be robust to various
motion patterns of pillar trajectories across time.
Thus our recurrent model can understand different object dynamics, and generalize to variable frame lengths during inference without retraining.
More experiments and analysis are provided in \autoref{tab:frame_length_generalization} and the ablation studies.

\subsection{Implementation Details}
We conduct 3D object detection within a wide range of 164$\times$164 meters ($m$) square zone, centering on the top LiDAR sensor. Point clouds inside this region are voxelized into 2D pillars with 0.32$m$ spatial resolutions.
The window attention blocks are based on 10$\times$10 grouping sizes.
The loss weights $\lambda_1$, $\lambda_2$ defined in (\ref{eq:total_loss}) are 200, 10 respectively.
We use AdamW \cite{kingma2014adam, loshchilov2017decoupled} optimizer with 128 batch sizes and 240$k$ iterations for distributed training on 128 TPUv3. The training takes about 2 days.
TPU memory usage is 5.4 GB on average and 7.4 GB at peak.
The first 10$k$ steps will warm up the learning rate from 5.0e-4 to 1.0e-3, after which the learning rate will follow a cosine annealing schedule to zero.
\begin{table}
\begin{center}
\caption{\textbf{Detection results on challenging large objects}.}
\resizebox{0.85\linewidth}{!}{%
\begin{tabular}{l|cc|cc}
\toprule
\multirow{2}{*}{Method}       &  \multicolumn{2}{c|}{L1} & \multicolumn{2}{c}{L2}  \\
&  2D & 3D & 2D & 3D  \\
\midrule
RSN \cite{sun2021rsn}   & \multicolumn{1}{c|}{53.10}                    & 45.20                   & \multicolumn{1}{c|}{-}                    & 40.90  \\
SWFormer \cite{sun2022swformer}   & \multicolumn{1}{c|}{58.33}                    & 49.74                 & \multicolumn{1}{c|}{53.45}                    & 45.23 \\
\textbf{LEF (ours)}  & \multicolumn{1}{c|}{\textbf{62.63}}                    & \textbf{54.35}                   & \multicolumn{1}{c|}{\textbf{57.42}}                    & \textbf{49.34}                   \\ \bottomrule
\end{tabular}%
}
\label{tab:main_results_val}
\end{center}
\vspace{-12 pt}
\end{table}

\section{Experiments}
\label{sec:exp}
In this section, we will compare our model with other state-of-the-art methods, and perform ablation studies upon the impact of our designs on detection performance.

\subsection{Dataset and Backbone}
We choose Waymo Open Dataset~\cite{sun2020scalability} over nuScenes~\cite{caesar2020nuscenes} and KITTI~\cite{geiger2012we} because WOD has large-scale and high-quality LiDAR data, which can better simulate the settings for developing on-road fully autonomous vehicles.
There are about 160$k$ annotated training frames in WOD but only around 30$k$ frames in nuScenes. 
As for per-frame point cloud densities, WOD is $\sim$200$k$ and nuScenes is $\sim$30$k$.
Therefore WOD is widely used in recent LiDAR-based methods: PV-RCNN(++), SST, RSN, SWFormer and so on~\cite{yin2021center, Fan2022EmbracingSS, sun2021rsn, sun2022swformer, shi2020pv, Shi2021PVRCNNPF, Ngiam2019StarNetTC, Chai2021ToTP, yang20213d, chen2022mppnet}.
WOD has 798 training sequences, 202 validation and 150 test sequences, covering diverse driving scenarios and agent status.
LiDAR data collection frequency is 10Hz.
Each frame of point clouds consists of data gathered from five sensors: one long-range and four short-range LiDAR.
For evaluation metrics, we adopt the officially recommended 3D AP / APH under two difficulty levels (L1, L2) depending on point densities of the ground-truth bounding boxes. APH is a weighted metric of AP using heading angles (\ie, azimuth).

We adopt the state-of-the-art SWFormer~\cite{sun2022swformer} as our detection backbone, and replace its original early-to-early LiDAR fusion with our proposed LEF. For fair comparisons, all training settings are kept the same as~\cite{sun2022swformer}.

\subsection{Main Results and Comparisons}
\label{sec:main_exps}

The overall vehicle detection results with other competing methods are in \autoref{tab:main_results_test}.
We compare against methods both with and without box refinement steps, although our model is a single-stage method without refinement and generally more efficient than those with box refinement.
Our method LEF surpasses the prior best single-stage model SWFormer by +1.3 3D APH on L2 test data (\eg 75.16 \vs\ 73.87), demonstrating the strong overall performance of our approach.

Our method is particularly useful for detecting challenging large objects whose maximum dimension is beyond 7 meters: truck, bus, construction vehicle, \etc\ We conduct detailed analysis on validation set in  \autoref{tab:main_results_val}.
Our method LEF outperforms SWFormer by +9.3\% relative increase on L1 3D AP: 54.35 \vs\ 49.74.
Hard cases such as large vehicles suffer from partial observation issues more often than small or medium size objects. Faithfully detecting these challenging cases requires LiDAR temporal fusion at long frame lengths in order to enlarge the sensory data coverage. Moreover, our late-to-early fusion scheme can reuse learned scene and object-aware latent features from prior frames, not simply stacking the point clouds as in RSN and SWFormer. Such high-level history knowledge can enable the model to more easily tackle challenging detection cases, compared with solving them from scratch using stacked raw sensory inputs.

\begin{figure}[t]
\centering
\includegraphics[width=1.0\linewidth]{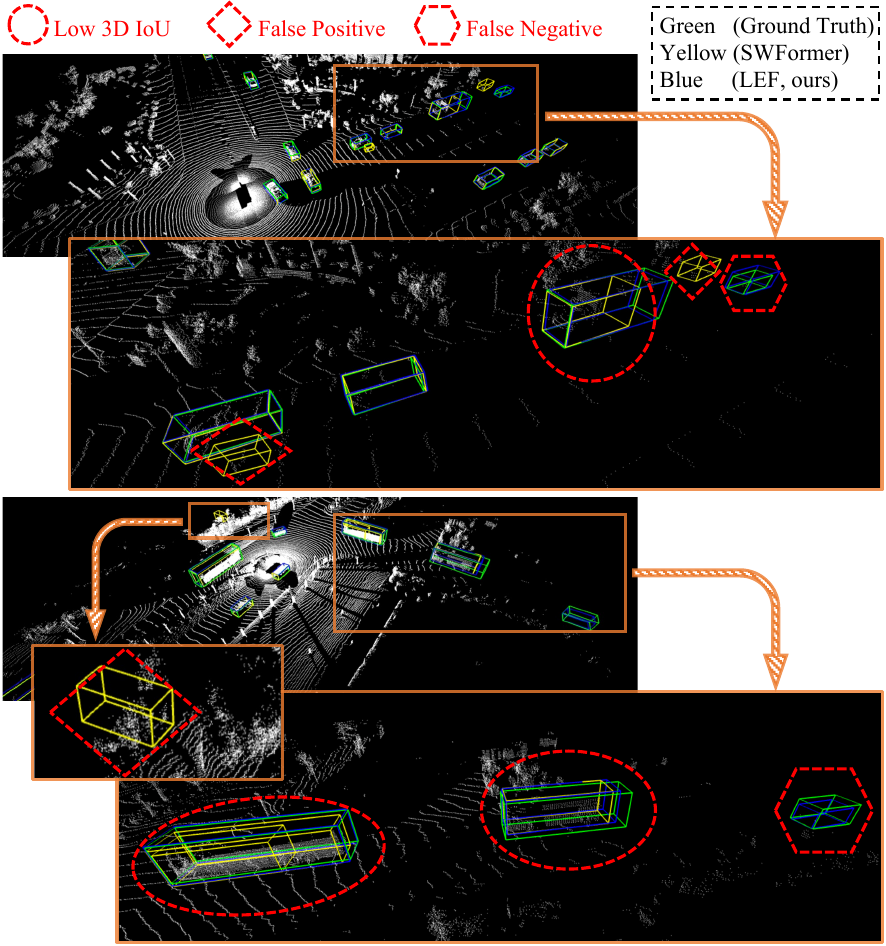}
\captionof{figure}{
Box colors are explain in the legend.
\textit{Errors of the baseline SWFormer} are highlighted in dashed red regions.}
\label{fig:visual_cmp}
\vspace{-13 pt}
\end{figure}

Qualitative results are visualized in \autoref{fig:visual_cmp}. Typical errors of SWFormer are highlighted in the red zones. Our results are aligned better (\ie, have higher 3D IoU) with the ground truth boxes than SWFormer predictions, especially for challenging large objects. Moreover, our results contain fewer false negative and false positive predictions than SWFormer results. We also measure model latency, flops and parameter sizes of different LiDAR 3D object detectors in \autoref{tab:latency_cmp}, following the same benchmark settings as~\cite{sun2022swformer}. PointPillars and SWFormer both use point stacking. The results demonstrate the efficiency advantages of our late-to-early recurrent fusion method.

\subsection{Ablation Studies}
\label{sec:ablation_studies}

\textbf{Fusion strategy.}
We conduct apple-to-apple comparisons to study the effect of early-to-early (E2E), late-to-late (L2L) and late-to-early (L2E) fusion strategies as illustrated in \autoref{fig:teaser_strategy}.
Specifically, we test all fusion variants with the same backbone and frame number (\ie, 3) to factorize out the influence of model architectures and LiDAR sequence lengths.
Results on validation set large objects are in \autoref{tab:fusion_type}.
Our L2E fusion surpasses the other two methods with 7.8\% relative gains on L1 3D AP. By comparing E2E and L2L fusion, we observe that their results on 2D AP are comparable. But E2E clearly outperforms L2L on 3D AP, indicating higher 3D object detection quality. 
These results validate our arguments about the benefits of late-to-early fusion. Compared with E2E fusion, L2E enables the model to reuse learned scene and object-aware knowledge from prior frames. Compared with L2L, the model capacity of L2E fusion is not constrained because its backbone has early access to the temporally aggregated sensory data.

\begin{table}[t]
\begin{center}
\caption{\textbf{Computation cost}. For fair comparisons, we use 3-frame temporal fusion settings on WOD for measurement.}
\resizebox{0.84 \linewidth}{!}{%
\begin{tabular}{l|c|c|c}
\toprule
Method              & Latency       & Flops        & Parameters        \\ \midrule
PointPillars~\cite{lang2019pointpillars}        & 93ms          & 375G         & 6.4M          \\
SWFormer~\cite{sun2022swformer}            & 47ms          & 35G          & 4.4M          \\
\textbf{LEF (ours)} & \textbf{38ms} & \textbf{29G} & 4.6M          \\ \bottomrule
\end{tabular}%
}
\label{tab:latency_cmp}
\end{center}
\vspace{-6 pt}
\end{table}

\begin{table}[!]
\begin{center}
\caption{\textbf{Ablation studies on different types of temporal fusion schemes}. All methods are trained with SLF.}
\resizebox{0.85\linewidth}{!}{%
\begin{tabular}{l|cc|cc}
\toprule
Fusion Strategy    & \multicolumn{2}{c|}{\begin{tabular}[c]{@{}c@{}}L1\\ 2D\ \ \ \ \ \ \ 3D\end{tabular}} & \multicolumn{2}{c}{\begin{tabular}[c]{@{}c@{}}L2\\ 2D\ \ \ \ \ \ \ 3D\end{tabular}} \\ \midrule
Early-to-Early & \multicolumn{1}{c|}{58.33}                    & 49.74                   & \multicolumn{1}{c|}{53.45}                    & 45.23                   \\
Late-to-Late   & \multicolumn{1}{c|}{58.74}                    & 48.83                   & \multicolumn{1}{c|}{53.67}                    & 44.32                   \\
\textbf{Late-to-Early}  & \multicolumn{1}{c|}{\textbf{61.46}}                    & \textbf{53.13}                   & \multicolumn{1}{c|}{\textbf{56.37}}                    & \textbf{48.28}                   \\ \bottomrule
\end{tabular}%
}
\label{tab:fusion_type}
\end{center}
\vspace{-5 pt}
\end{table}

\begin{table}[!]
\begin{center}
\caption{\textbf{Ablation studies on different object sizes}. The 3D AP gains achieved by LEF increase as object size grows.}
\resizebox{1.0\linewidth}{!}{
\begin{tabular}{l|ccc|ccc}
\toprule
\multirow{2}{*}{Method} & \multicolumn{3}{c|}{L1}                                         & \multicolumn{3}{c}{L2}                                          \\
& Large     & Medium     & Small & Large     & Medium    & Small \\ \midrule
RSN \cite{sun2021rsn}                & \multicolumn{1}{c|}{45.20} & \multicolumn{1}{c|}{77.30} & 79.40 & \multicolumn{1}{c|}{40.90} & \multicolumn{1}{c|}{68.60} & 69.90 \\
SWFormer \cite{sun2022swformer}                & \multicolumn{1}{c|}{49.74} & \multicolumn{1}{c|}{79.11} & 82.36 & \multicolumn{1}{c|}{45.23} & \multicolumn{1}{c|}{70.59} & 74.04 \\
\textbf{LEF (ours)}              & \multicolumn{1}{c|}{\textbf{54.35}} & \multicolumn{1}{c|}{\textbf{79.62}} & \textbf{82.46} & \multicolumn{1}{c|}{\textbf{49.34}} & \multicolumn{1}{c|}{\textbf{71.32}} & \textbf{74.15} \\ \bottomrule
\end{tabular}%
}
\label{tab:object_size}
\end{center}
\vspace{-16 pt}
\end{table}

\textbf{Different object sizes.}
Besides the overall results and hard example analysis in Section \ref{sec:main_exps}, we are also interested in knowing the impact of our method on different object sizes.
Thus we divide validation set objects into: large, medium and small. Typical large objects are bus and truck. Medium and small objects usually include sedan and pedestrian, respectively. Detailed results are in \autoref{tab:object_size}.
Although our method LEF achieves comparable results with the competing methods on small objects, we observe increasingly more gains as object sizes grow larger. On L2 medium objects, LEF improves SWFormer by 0.73 AP and the gains further bump to 4.11 AP on large objects.
One possible explanation is that small objects suffer less from partial-view observation issues than large objects, and thus do not significantly benefit from temporal fusion.
From the results we believe that our method works robustly across different object sizes.

\textbf{Frame length generalization.}
Due to memory constraint of the computing devices, GPU or TPU, 3D object detectors with LiDAR temporal fusion usually sample a fixed number of history frames (\eg 2 or 3) during training. However, during inference, there are usually additional frames available to the model depending on the history lengths. For typical early-to-early fusion based multi-frame detectors (\eg CenterPoint, SWFormer), if we want to test a trained model on different frame lengths, the training settings need to be modified and the model needs to be retrained.
With stochastic-length FrameDrop (SLF), LEF can generalize to variable frame lengths \textit{without} retraining.
It can leverage additional frames and achieve increasingly improved results.
Large objects 3D AP are shown in \autoref{tab:frame_length_generalization}.
In contrast, SWFormer and LEF without SLF can not make best of long history and might even face performance decrease. This is because long history frames can exhibit diverse motion patterns of temporally aggregated data, posing generalization difficulties for methods trained without SLF. Moreover, since SWFormer is based on point cloud stacking, it will run into OOM if we simply stack
a long LiDAR sequence into millions of 3D points
and use them as inputs. These observations indicate that stochastic-length FrameDrop and recurrent fusion are critical in generalizing our method LEF to variable frame lengths during inference.


\textbf{Foreground pillar segmentation.} To efficiently fuse history pillar features in a recurrent manner, we apply BEV foreground segmentation before passing history latent pillar embeddings into the next frame.
the number of history pillars that need to be recurrently fused can be reduced from $\sim$20$k$ to $\sim$2$k$ on average after removing a huge amount of uninformative background data.
Therefore the computation burden of our late-to-early temporal fusion scheme can be greatly reduced and maintained at a relatively low constant cost.

\begin{table}[t]
\begin{center}
\caption{\textbf{Long frame history generalization studies}. For each trained model, we evaluate its inference generalization ability to different frame (f) lengths \textit{without} retraining.
}
\resizebox{1.0\linewidth}{!}{
\begin{tabular}{l|ccc|ccc}
\toprule
\multirow{2}{*}{Method} & \multicolumn{3}{c|}{L1}                                         & \multicolumn{3}{c}{L2}                                          \\
& 3-f     & 6-f     & 9-f & 3-f     & 6-f    & 9-f \\ \midrule
SWFormer \cite{sun2022swformer}                & \multicolumn{1}{c|}{46.23} & \multicolumn{1}{c|}{38.76} & OOM       & \multicolumn{1}{c|}{41.93} & \multicolumn{1}{c|}{35.09} & OOM       \\
LEF (w/o SLF)                & \multicolumn{1}{c|}{51.18} & \multicolumn{1}{c|}{51.44} & 50.84       & \multicolumn{1}{c|}{46.58} & \multicolumn{1}{c|}{46.91} & 46.28       \\
\textbf{LEF (with SLF)}              & \multicolumn{1}{c|}{\textbf{53.13}} & \multicolumn{1}{c|}{\textbf{53.96}} & \textbf{54.35} & \multicolumn{1}{c|}{\textbf{48.28}} & \multicolumn{1}{c|}{\textbf{48.99}} & \textbf{49.34} \\ \bottomrule
\end{tabular}%
}
\label{tab:frame_length_generalization}
\end{center}
\vspace{-5 pt}
\end{table}

\begin{table}[]
\begin{center}
\caption{\textbf{Inverse calibration and alignment (ICA)} can improve detection AP across different object sizes.}
\resizebox{0.92\linewidth}{!}{
\begin{tabular}{l|cc|cc|cc}
\toprule
\multirow{2}{*}{ICA} & \multicolumn{2}{c|}{Large}         & \multicolumn{2}{c|}{Medium}        & \multicolumn{2}{c}{Small}          \\
                        & 2D                         & 3D    & 2D                         & 3D    & 2D                         & 3D    \\ \midrule
w/o                     & \multicolumn{1}{c|}{60.85} & 51.34 & \multicolumn{1}{c|}{92.72} & 78.30 & \multicolumn{1}{c|}{85.92} & 80.59 \\
\textbf{with}                    & \multicolumn{1}{c|}{\textbf{62.63}} & \textbf{54.35} & \multicolumn{1}{c|}{\textbf{93.02}} & \textbf{79.62} & \multicolumn{1}{c|}{\textbf{87.40}} & \textbf{82.46} \\ \bottomrule
\end{tabular}%
}
\label{tab:temporal_alignment}
\vspace{-14pt}
\end{center}
\end{table}

\textbf{Inverse calibration and alignment.}
Inverse calibration and alignment, as illustrated in \autoref{fig:pipeline}, is important for fusing two sparse sets of pillar features between the prior and the current frames. Features belonging to the same underlying static objects can be effectively aligned after this temporal alignment process.
In \autoref{tab:temporal_alignment} we show that inverse calibration and alignment achieves consistent detection improvement across different size objects, including truck, sedan, pedestrian, and so on.

\begin{table}[t]
\begin{center}
\caption{\textbf{Variants of window-based attention blocks for recurrent temporal fusion}.
Based on the comparisons, we adopt self-attention as default in other experiments.}
\resizebox{0.825\linewidth}{!}{
\begin{tabular}{l|cc|cc}
\toprule
\multirow{2}{*}{Attention Type} & \multicolumn{2}{c|}{L1}          & \multicolumn{2}{c}{L2}    \\
 & 2D & 3D & 2D & 3D \\
\midrule
Cross-Attn   & 51.69 & 42.35 & 47.06 & 38.36 \\
Mix-Attn      & 61.68 & 52.94 & 56.46 & 48.06 \\
\textbf{Self-Attn}   & \textbf{62.63} & \textbf{54.35} & \textbf{57.42} & \textbf{49.34} \\ \bottomrule
\end{tabular}%
}
\label{tab:fusion_ops}
\end{center}
\vspace{-3pt}
\end{table}

\begin{table}[]
\begin{center}
\caption{\textbf{The impact of window-based self-attention on different speed objects}.} 
\resizebox{1.0 \linewidth}{!}{%
\begin{tabular}{l||c|c|c||c|c}
\toprule
Self-Attention & Static        & Slow          & Medium        & Fast          & Very Fast     \\ \midrule
without      & 60.55         & 63.46         & 74.58         & 53.07         & 75.47         \\
\textbf{with}     & \textbf{66.62} & \textbf{69.27} & \textbf{79.62} & \textbf{62.46} & \textbf{82.14} \\ \bottomrule
\end{tabular}%
}
\label{tab:with_without_attention}
\end{center}
\vspace{-14pt}
\end{table}

\textbf{Window-based Attention Fusion.} We apply window-based attention blocks on temporally aligned sparse pillar tokens to further fuse information of the history and current frames. As explained in Section \ref{sec:method_window_attention}, we explore three different attention designs: self / cross / mix-attention.
Detection AP on large objects of WOD validation set are shown in \autoref{tab:fusion_ops}.
For all methods, we use the sparse set of pillar tokens $\{ V^{''}_{i,u} \}$ converted from the temporally aligned BEV feature map $J^{'}_{i}$ as the query tensor.
In self-attention, query, key and value are based on the same tensor.
In cross-attention, the key and value tensors are the sparse set of pillar tokens $\{ \tilde{V}_{i-1,c} \}$ converted from the calibrated history features $\tilde{I}_{i-1}$.
Mix-attention uses the union set of prior methods as key and value.
We observe that self-attention consistently outperforms the other two attention variants. This is presumably because the history tokens exist in a quite different latent space from the temporally aligned tokens. Therefore attention between $\{ \tilde{V}_{i-1,c} \}$ and $\{ V^{''}_{i,u} \}$ might easily lead to intractable feature fusion and eventually hurt detection. Meanwhile, since $J^{'}_{i}$ has already merged information from the history $\tilde{I}_{i-1}$ and the current $I_{i}$, self-attention is competent to associate relevant pillar tokens and fulfill the fusion task.

Window-based attention fusion plays an important role in fusing the information from moving object pillars.
In \autoref{tab:with_without_attention}, we present validation set 3D AP comparisons between with and without window-based self-attention fusion.
We report subcategory metrics under different speed ranges:
[0, 0.45), [0.45, 2.24), [2.24, 6.71), [6.71, 22.37), [22.37, +$\infty$) miles per hour for static, slow, medium, fast, very fast objects.
The metrics are averaged over different size objects.
We observe that attention fusion brings consistent detection gains across different object speed ranges. Particularly, the improvements achieved on high-speed objects are larger than those on low-speed objects: +9.4 (fast) \vs\ +6.1 (static) 3D AP gains. The comparisons empirically prove that window-based self-attention fusion is critical in associating relevant pillars that belong to the same underlying objects, which is especially important for moving object detection.

\section{Conclusions and Future Work}
\label{sec:conclusion}
In this paper, we conduct an in-depth study on the temporal fusion aspect of 3D object detection from LiDAR sequences.
We propose a late-to-early temporal feature fusion method that recurrently extracts sparse pillar features from both object-aware latent embeddings and LiDAR sensor raw inputs.
To handle the alignment issues of static and moving objects, we propose inverse calibration and alignment as well as window-based attention fusion methods.
We also apply foreground segmentation to obtain sparse pillar features from history for computation reduction.
The resulting model, LEF, performs favorably against its base model SWFormer in both detection quality and efficiency. The improvement is especially significant on large objects that require multiple LiDAR sweeps fused across space and time to achieve high surface coverage rate.

As future work, we plan to extend our method to multi-modal sensor fusion with a focus on integrating camera and radar information.
Recurrent late-to-early temporal fusion schemes like ours and BEVFormer~\cite{li2022bevformer} have been explored in very few papers. To further demonstrate the effectiveness of this approach, it would be beneficial to test it on various backbone models and extend its application beyond the scope of 3D object detection task.

\bibliographystyle{./IEEEtran} 
\bibliography{./IEEEabrv,./IEEEexample}

\end{document}